# MAJORScore: A Novel Metric for Evaluating Multimodal Relevance via Joint Representation

Zhicheng Du[1,2,*,†], Qingyang Shi[1,2,*], Jiasheng Lu[2], Yingshan Liang[1,2],
Xinyu Zhang[1,2], Yiran Wang[1,2], Peiwu Qin[1,†]
[1]Shenzhen International Graduate School, Tsinghua University, Shenzhen, China
[2]Huawei Technologies Co., Ltd., Shenzhen, China

*Abstract*—The multimodal relevance metric is usually borrowed from the embedding ability of pretrained contrastive learning models for bimodal data, which is used to evaluate the correlation between cross-modal data (e.g., CLIP). However, the commonly used evaluation metrics are only suitable for the associated analysis between two modalities, which greatly limits the evaluation of multimodal similarity. Herein, we propose MAJORScore, a brand-new evaluation metric for the relevance of multiple modalities (N modalities, N>=3) via multimodal joint representation for the first time. The ability of multimodal joint representation to integrate multiple modalities into the same latent space can accurately represent different modalities at one scale, providing support for fair relevance scoring. Extensive experiments have shown that MAJORScore increases by 26.03%-64.29% for consistent modality and decreases by 13.28%-20.54% for inconsistence compared to existing methods. MAJORScore serves as a more reliable metric for evaluating similarity on large-scale multimodal datasets and multimodal model performance evaluation.

## I. INTRODUCTION

The development of large-scale multimodal datasets and benchmarks has surged, driven by the need to train and evaluate generative models that leverage diverse data sources [1]–[3]. As each modality provides a distinct perspective on the same content, the number of modalities within these datasets has expanded from two to three or more [4]–[7]. This rapid growth, encompassing modalities such as text, audio, vision (image and video), and time-series data, has highlighted the urgent need for reliable and equitable methods to assess the relevance and consistency of information across different modalities [8], [9]. Moreover, research has shown that generative models utilizing multimodal conditional controls yield superior results compared to those relying on single-modal inputs, prompting researchers to build large-scale multimodal datasets tailored to specific tasks and to design multimodal joint-condition generation models [10]–[12]. However, traditional evaluation metrics often struggle to capture the complex relationships among these diverse data types, leading to suboptimal performance in tasks such as information retrieval, recommendation systems, multimodal feature fusion, and content generation [13]–[16]. Hence, the absence of effective multimodal correlation metrics poses a significant challenge, hindering the development of high-consistency multimodal datasets and impeding advancements in multimodal generative model research.

TABLE I
EXAMPLE SAMPLE PAIRS OF CLIP AND CLAP SCORING SELECTED FROM VGGSOUND DATASET.

| Sample ID | Text | CLIP | CLAP | Diff. |
|---|---|---|---|---|
| AD3yJE3A2eY 70 | horse neighing | 0.2994 | -0.4309 | 0.7303 |
| Tx1k80M1dGI 0 | wind chime | 0.1547 | 0.7695 | 0.6148 |

Diff. means difference.
Details of video and audio can be accessed at project website.

The commonly used metrics for assessing multimodal similarity are mainly borrowed from pretrained contrastive learning models, such as Contrastive Language-Image PreTraining (CLIP) [17] and Contrastive Language-Audio Pretraining (CLAP) [18]. These models generate feature embeddings for images and text from aligned CLIP encoders, and the correlation between visual and textual data is assessed by measuring the similarity between their embeddings. Consistency evaluation across three or more modalities typically involves combining the results of contrastive models trained on different bimodal data (e.g., integrating CLIP and CLAP). Vision-text and audio-text similarity scores are obtained using CLIP's image-text encoders and CLAP's audio-text encoders, respectively, and these scores are then combined to produce a final consistency score for the vision-text-audio trimodality [10]. Another method [19] computes the distance between audio embeddings and video embeddings from VGGish [20] and text embeddings from word2vec [21]. However, we observe that differences between the embedding spaces of the two contrastive models result in an imbalance between vision-text and audio-text scores, making the final consistency score unreliable and biased (Table I). This discovery proves that the straightforward concatenation of contrast pretraining models embedded by divergent encoding spaces for the purpose of multimodal correlation assessment is not a tenable approach. A more viable strategy involves assessing similarity by integrating the encoding of multiple modalities into a shared latent space. In addition to the above methods and relatively expensive subjective testing [22], [23], Diff-Foley [24] is evaluated by training an alignment classifier to access the alignment degree of input video and generated audio. Nevertheless, due to the absence of reliable multimodal correlation metrics, many studies are still limited to qualitative analyses without quantitative comparisons for vision-text-audio relevance [23], [25], [26]. In a nutshell, current evaluation metrics for multimodal relevance inadequately encompass the intricate interconnections across modalities, rendering the evaluation of multimodal relevance a continual challenge.

Recent advances in deep learning and representation learning have paved the way for more effective integration of multimodal data through joint representation learning. Given the fundamental role of multi-modal joint representations in understanding and generation pipelines, a high-quality full joint representation would be a step towards collaborative processing of more diverse multi-modal information. A key challenge in this context is the development of a unified latent space that can encode diverse modal data, providing the foundation for reliable and fair multimodal similarity assessment. In this paper, we design a novel metric called **MAJORScore** (**M**ultimod**A**l **J**oint **R**epresentation Score), designed to evaluate multimodal correlations through joint representation learning. MAJORScore uses a unified latent space to encode different modalities, facilitating the measurement of inter-modality correlations on a common scale. To the best of our knowledge, this is the first work to employ multimodal joint representation learning for the task of assessing multimodal consistency. We deployed extensive experiments to evaluate the ability of MAJORScore to evaluate multimodal data pairs. According to different application scenarios, we evaluate the generalization ability of MAJORScore on various datasets and provide detailed analysis based on the experimental results. This paper is organized as follows: Section II describes the design and calculation method of MAJORScore. Section III depicts the experimental design and experimental results. Section IV provides a discussion and conclusion.

## II. MAJORScore

In this section, we outline the methodology for calculating the MAJORScore in detail. The proposed approach consists of three primary phases: (i) extracting embeddings for each modality (i.e., video, text, and audio), (ii) computing the similarity scores between video-text and text-audio pairs, and (iii) deriving the MAJORScore via calculating the aggregation features of the similarity scores. The complete pipeline of computing MAJORScore is shown in Figure 1.

### A. Multimodal joint representation

The ability of multimodal joint representation to encode multiple modalities in a powerful and unified way is not only useful in large foundation models, but we find that this ability can also be applied to multi-modal associativity ($N >= 3$) for reliable evaluation. Based on this, we propose the first multimodal relevance and similarity fairness evaluation metric based on multimodal joint representation. To demonstrate the effectiveness of MAJORScore, we employ Connecting Multimodal Contrastive Representation (C-MCR) [27] as the multimodal joint representation model to evaluate the relevance of video-text-audio.C-MCR integrates the representations from CLIP and CLAP using textual connections, thereby generating comprehensive audio-visual representations. This approach, devoid of the need for paired data or fine-tuning, has demonstrated superior performance on six distinct datasets, excelling in three key downstream audiovisual tasks.

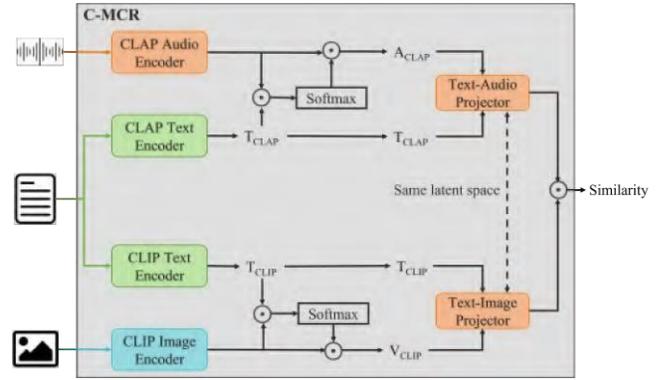

Fig. 1. Pipeline of MAJORScore. The C-MCR model is one of the representative multimodal joint representation models that binds the latent spaces of CLIP and CLAP to the same space.

---

**Algorithm 1** Computing MAJORScore

1: **Input:** Vision data $V$, Text data $T$, Audio data $A$
2: **Output:** $MAJORScore_{sum}$, $MAJORScore_{prod}$, $MAJORScore_{avg}$
3: **Step 1:** Extracting embeddings from C-MCR model
4: $E_V \leftarrow$ C-MCR vision encoder$(V)$
5: $E_T \leftarrow$ C-MCR text encoder$(T)$
6: $E_A \leftarrow$ C-MCR audio encoder$(A)$
7: **Step 2:** Computing Cosine Similarities
8: $S_{VT} \leftarrow$ CosineSimilarity$(E_V, E_T)$
9: $S_{TA} \leftarrow$ CosineSimilarity$(E_T, E_A)$
10: **Step 3:** Compute Sum, Product, and Average
11: $MAJORScore_{sum} \leftarrow S_{VT} + S_{TA}$
12: $MAJORScore_{prod} \leftarrow S_{VT} * S_{TA}$
13: $MAJORScore_{avg} \leftarrow \frac{1}{2}(S_{VT} + S_{TA})$
14: **Return:** $MAJORScore_{sum}$, $MAJORScore_{prod}$, $MAJORScore_{avg}$

---

### B. Computing MAJORScore

The multimodal joint representation model provides fundamental support for fairly measuring the relevance between different modalities in a unified embedding encoding space [28]. First, we encode data from the three modalities—video, text, and audio—using aligned encoders based on the C-MCR model, resulting in visual embeddings ($V_{emb}$), text embeddings

($T_{emb}$), and audio embeddings ($A_{emb}$) in a unified latent space. Specifically, we follow the workflow similar to previous works [23], [29] of using CLIP to extract the features of video. The video is extracted at one frame per second to get the image sequence, then encoded using CLIP, and finally averaged to obtain the embeddings. Given that video and audio frequently co-occur in real-world scenarios, and to ensure fair comparison with previous evaluation methods, we compute the cosine similarity between vision-text and text-audio pairs as follows:

$$CMCR_{vt} = cos(V_{emb}, T_{emb}) \quad (1)$$

$$CMCR_{ta} = cos(T_{emb}, A_{emb}) \quad (2)$$

In order to comprehensively compare the evaluation accuracy of the metrics on the sample pairs, we used the method of product, sum and average to synthesize the absolute value of the two cosine similarities. The specific calculation formula is:

$$MAJORScore = f(CMCR_{ta}, CMCR_{vt}) \quad (3)$$

where $f()$ means the function to combine two similarities, including product, sum, and average. Additionally, to measure multimodal association more comprehensively, we introduce the concept of similarity fairness, which indicates the degree of balance in the consistency across modalities. A smaller difference between multiple similarity scores implies better balance and prevents the high correlation of a single modality from disproportionately influencing the overall similarity. If similarity fairness of $M$ modalities are evaluated, $C_M^2$ bimodal similarity scores are computed at first and the formula is expressed as:

$$FairScore = \frac{1}{C_M^2} \sum_{1 \leq i < j \leq M} |S_i - S_j| \quad (4)$$

where $S$ means bimodal similarity score. Algorithm 1 summarize the calculation process of MAJORScore, where prod and avg are the abbreviation of product and average.

### C. Baseline method

As a baseline, we adopt the metrics from recent work on multimodal conditional control generative models that evaluate image-text-audio similarity [10]. The baseline method uses the CLIP and CLAP models to compute image-text similarity and text-audio similarity, respectively, expressed as follows:

$$CLAPScore = cos(T_{emb}, A_{emb}) \quad (5)$$

$$CLIPScore = cos(T_{emb}, V_{emb}) \quad (6)$$

We adopt the same method to comprehensively consider two cosine similarity, and the specific calculation formula is:

$$CLIPCLAP = f(CLAPScore, CLIPScore) \quad (7)$$

where the function $f()$ has the same meaning as in the MAJORScore calculation in section II-B.

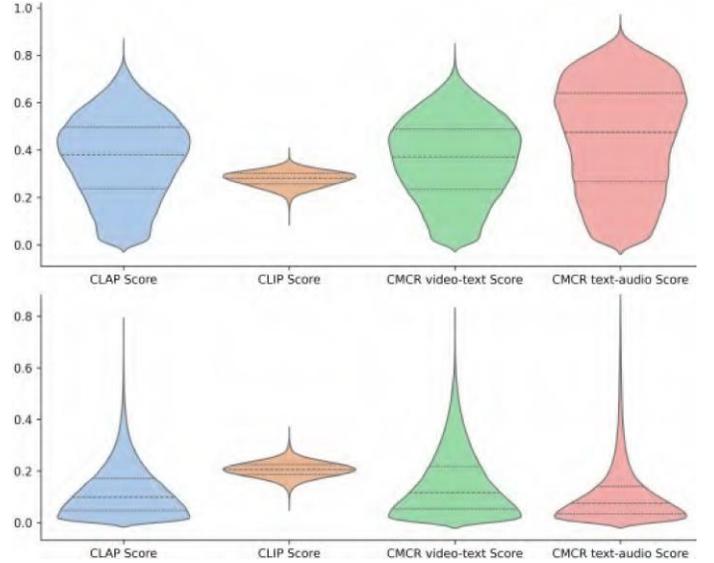

Fig. 2. Cross-modal similarity scoring results from CLIP, CLAP, and C-MCR for vision-text and text-audio in VGGSound datasets. The picture above is comparisons for the consistent case of modal data, and the picture below is for the inconsistent case.

## III. EXPERIMENTS

### A. Datasets and data processing

We conduct the vision-text-audio modal similarity evaluation experiments on the VGGSound dataset [2] and a newly collected dataset consisting of multi-modal inputs and audio outputs from multiple audio generation models. The VGGSound dataset is a comprehensive audio-visual dataset designed for training and evaluating audio recognition models. It consists of approximately 200K 10-second short video-audio clips extracted from videos uploaded to YouTube, ensuring a diverse range of real-world acoustic environments and noise characteristics. Considering that text is an intermediate modality connecting video and audio and the inherent high correlation of real-world video and audio, we built on top of the VGGSound dataset with modality mismatch sample pairs (consistent vision-audio modality with mispaired text) as the negative samples for the experiment, named VGGSound mispaired dataset. Furthermore, based on the visual-text input and audio output of the Foley-Crafter [26], SVA [25] and seeing-and-hearing models [30], 37 visual-text-audio sample pairs are constructed as a Visual-Text-Audio Synthesis (VITAS) dataset to access the performance of MAJORScore in generative model performance evaluation.

### B. Implementation Details

We evaluate the performance of MAJORScore on the modal consistent, modal inconsistent, and multimodal conditional generated content tasks in the VGGSound, VGGSound mispaired dataset, and VITAS dataset, respectively. As a comparison, we use the evaluation method of fusing CLIP and CLAP used in previous works as baseline. The comparative test was divided into quantitative analysis and qualitative analysis. The quantitative analysis experiment consists of two parts:

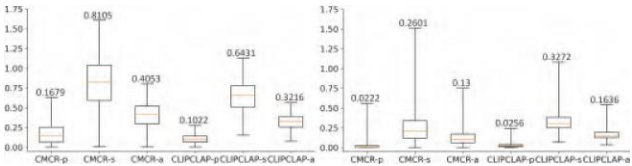

Fig. 3. Multimodal relevance results comparison. The picture above is results of the consistent case of modal data, and the picture below is for the inconsistence. $p$ means product, $s$ represents sum and $a$ is average.

the bimodal similarity fairness and the trimodal relevance scores. The evaluation criterion of the experiment is that the closer the bimodal similarity score represents, the better fairness, and the trimodal correlation score of vision-text-audio should be higher when the modalities are consistent and lower when they are inconsistent. The qualitative analysis experiment refers to selecting sample pairs with relatively low multimodal relevance scores in the VGGSound dataset and VITAS dataset for content assessment and analyzing whether the reasons for low scores are related to modal content inconsistency.

TABLE II
CORRELATION COEFFICIENT ANALYSIS OF DATA DISTRIBUTION FROM CLIP, CLAP, AND C-MCR FOR VISION-TEXT AND TEXT-AUDIO SCORES OF THE VGGSOUND DATASET.

| Metric | $CMCR_{vt}$ and $CMCR_{ta}$ | | $CLIP$ and $CLAP$ | |
|---|---|---|---|---|
| Case | Consistent | Inconsistent | Consistent | Inconsistent |
| Mean Diff↓ | 0.0933 | 0.0395 | 0.0869 | 0.0845 |
| Cohen's d↓ | 0.4684 | 0.3278 | 0.6959 | 1.162 |
| T-Test Value↓ | 147.8 | 103.5 | 219.6 | 366.8 |
| Std Dev Diff↓ | 0.0575 | 0.0071 | 0.1403 | 0.0679 |
| Skewness↓ | 0.0371 | 1.200 | 0.2136 | 1.282 |

### C. Quantitative Comparison

The experimental outcomes pertaining to bimodal similarity fairness and trimodal relevance scores are delineated in Figure 2 and Figure 3, respectively. Figure 2 implies that the distributions of video-text and text-audio scores for the CMCR model exhibit greater proximity than those of intergating CLIP and CLAP. The variability between video-text and text-audio similarity scores is quantified using multiple metrics, as presented in Table II. The term "Mean Diff" refers to the mean difference, which quantifies the average disparity between two sets of values. "Std Dev Diff" denotes the standard deviation, a measure of the variability within a dataset. Cohen's d is employed to assess the extent of the difference between two groups, while the T-Test Value ascertains whether a significant divergence exists between the means of two groups. Skewness indicates the degree of asymmetry in the distribution of a random variable around its mean. The findings depicts that (i) MAJORScore effectively gauges the divergence in cross-modal correlations of incongruent data on a uniform scale (Figure 2), and (ii) MAJORScore assigns higher scores (increasing 26.03%-64.29%) to sample pairs with congruent modalities and lower scores (decreasing 13.28%-20.54%) to those with inconsistent sample pairs, surpassing the performance of baseline methods (Figure 3).

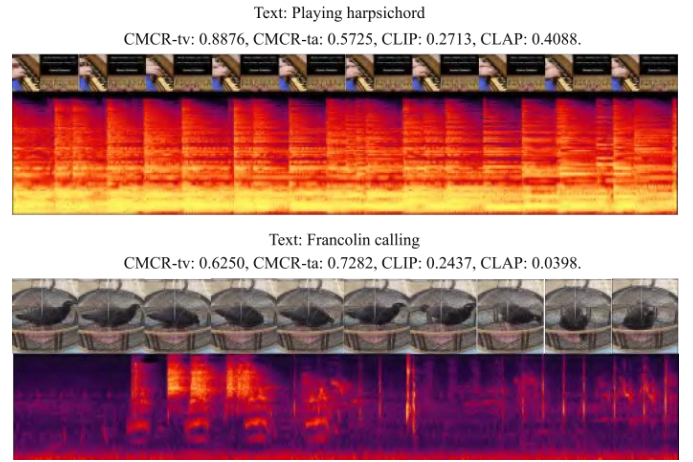

Fig. 4. Qualitative comparison results from the VGGSound and VITAS datasets. The top picture reflects the high MAJORScore for the image-text case. The below figure shows the high MAJORScore for the text-audio case.

### D. Qualitative Comparison

Figure 4 illustrates examples with high and low MAJORScore in the VGGSound and VITAS datasets. The experimental results demonstrate that MAJORScore effectively identifies samples from both real-world and synthetic datasets that exhibit poor consistency. This capability suggests that MAJORScore can be utilized to (i) refine existing large-scale datasets, thereby enhancing the performance of multi-dimensional large models [31], [32], and (ii) rank and recommend multiple candidate content generated by models, filtering out synthetic data with low correlation [33].

## IV. DISCUSSION AND CONCLUSION

In this paper, we introduce an innovative metric termed MAJORScore, designed to assess multimodal relevance through joint representation learning for the first time. This method capitalizes on the intrinsic intermodal relationships to provide a more accurate correlation assessment. By amalgamating the strengths of each modality into a unified representation, MAJORScore enhances the assessment of multimodal correlations and provides new insights into the intrinsic relationships between disparate data types. Comprehensive experiments on video-text-audio datasets demonstrate the efficacy of MAJORScore, highlighting its potential to propel advancements in multimodal analysis and its applicability across various domains. Future investigations should prioritize multimodal consistency evaluation across a broader range of modalities. Notably, models such as Ex-MCR [34], Omnibind [16], [35], Imagebind [36], Languagebind [37], and Meta-transformer [38], which can embed an increasing number of modalities into the same latent space. The exploration of similarity across an extended spectrum of modalities, facilitated by these sophisticated multimodal joint representational methods, represents a promising trajectory for subsequent scholarly pursuits.

## References

[1] Jiaxi Gu, Xiaojun Meng, Guansong Lu, Lu Hou, Niu Minzhe, Xiaodan Liang, Lewei Yao, Runhui Huang, Wei Zhang, Xin Jiang, Chunjing XU, and Hang Xu, "Wukong: A 100 million large-scale chinese cross-modal pre-training benchmark," in *Advances in Neural Information Processing Systems*, S. Koyejo, S. Mohamed, A. Agarwal, D. Belgrave, K. Cho, and A. Oh, Eds. 2022, vol. 35, pp. 26418–26431, Curran Associates, Inc.

[2] Honglie Chen, Weidi Xie, Andrea Vedaldi, and Andrew Zisserman, "Vggsound: A large-scale audio-visual dataset," in *ICASSP 2020 - 2020 IEEE International Conference on Acoustics, Speech and Signal Processing (ICASSP)*, 2020, pp. 721–725.

[3] Jian Li, Weiheng Lu, Hao Fei, Meng Luo, Ming Dai, Min Xia, Yizhang Jin, Zhenye Gan, Ding Qi, Chaoyou Fu, Ying Tai, Wankou Yang, Yabiao Wang, and Chengjie Wang, "A survey on benchmarks of multimodal large language models," 2024.

[4] Rodrigo Benenson and Vittorio Ferrari, "From coulouring-in to pointillism: revisiting semantic segmentation supervision," in *ArXiv*, 2022.

[5] Jordi Pont-Tuset, Jasper Uijlings, Soravit Changpinyo, Radu Soricut, and Vittorio Ferrari, "Connecting vision and language with localized narratives," in *ECCV*, 2020.

[6] Rodrigo Benenson, Stefan Popov, and Vittorio Ferrari, "Large-scale interactive object segmentation with human annotators," in *CVPR*, 2019.

[7] Alina Kuznetsova, Hassan Rom, Neil Alldrin, Jasper Uijlings, Ivan Krasin, Jordi Pont-Tuset, Shahab Kamali, Stefan Popov, Matteo Malloci, Alexander Kolesnikov, Tom Duerig, and Vittorio Ferrari, "The open images dataset v4: Unified image classification, object detection, and visual relationship detection at scale," *IJCV*, 2020.

[8] Xiao Dong, Xunlin Zhan, Yangxin Wu, Yunchao Wei, Michael C. Kampffmeyer, Xiaoyong Wei, Minlong Lu, Yaowei Wang, and Xiaodan Liang, "M5product: Self-harmonized contrastive learning for e-commercial multi-modal pretraining," in *Proceedings of the IEEE/CVF Conference on Computer Vision and Pattern Recognition (CVPR)*, June 2022, pp. 21252–21262.

[9] Zhicheng Du, Zhaotian Xie, Yan Tong, and Peiwu Qin, "LAMPER: language model and prompt engineering for zero-shot time series classification," in *The Second Tiny Papers Track at ICLR 2024, Tiny Papers @ ICLR 2024, Vienna, Austria, May 11, 2024*. 2024, OpenReview.net.

[10] Sanjoy Chowdhury, Sayan Nag, Joseph K J, Balaji Vasan Srinivasan, and Dinesh Manocha, "Melfusion: Synthesizing music from image and language cues using diffusion models," *CVPR*, 2024.

[11] Yang Liu, Xiaoyun Zhong, Shiyao Zhai, Zhicheng Du, Zhenyuan Gao, Qiming Huang, Can Yang Zhang, Bin Jiang, Vijay Kumar Pandey, Sanyang Han, Runming Wang, Yuxing Han, Chuhui Wang, and Peiwu Qin, "Prompt-enhanced hierarchical transformer elevating cardiopulmonary resuscitation instruction via temporal action segmentation," *Computers in Biology and Medicine*, vol. 167, pp. 107672, 2023.

[12] Noah Cohen Kalafut, Xiang Huang, and Daifeng Wang, "Joint variational autoencoders for multimodal imputation and embedding," *Nature Machine Intelligence*, vol. 5, no. 6, pp. 631–642, 2023.

[13] Zhicheng Du, Chenyao Jiang, Xi Yuan, Shiyao Zhai, Zhengyang Lei, Shuyue Ma, Yang Liu, Qihui Ye, Chufan Xiao, Qiming Huang, Ming Xu, Dongmei Yu, and Peiwu Qin, "Game: Generalized deep learning model towards multimodal data integration for early screening of adolescent mental disorders," 2023.

[14] Meng Jiang, Alex Beutel, Peng Cui, Bryan Hooi, Shiqiang Yang, and Christos Faloutsos, "A general suspiciousness metric for dense blocks in multimodal data," in *2015 IEEE International Conference on Data Mining*, 2015, pp. 781–786.

[15] Zhicheng Du, Zhaotian Xie, Huazhang Ying, Likun Zhang, and Peiwu Qin, "Cognitive resilience: Unraveling the proficiency of image-captioning models to interpret masked visual content," in *The Second Tiny Papers Track at ICLR 2024, Tiny Papers @ ICLR 2024, Vienna, Austria, May 11, 2024*. 2024, OpenReview.net.

[16] Zehan Wang, Ziang Zhang, Hang Zhang, Luping Liu, Rongjie Huang, Xize Cheng, Hengshuang Zhao, and Zhou Zhao, "Omnibind: Large-scale omni multimodal representation via binding spaces," 2024.

[17] Alec Radford, Jong Wook Kim, Chris Hallacy, Aditya Ramesh, Gabriel Goh, Sandhini Agarwal, Girish Sastry, Amanda Askell, Pamela Mishkin, Jack Clark, Gretchen Krueger, and Ilya Sutskever, "Learning transferable visual models from natural language supervision," in *Proceedings of the 38th International Conference on Machine Learning*, Marina Meila and Tong Zhang, Eds. 18–24 Jul 2021, vol. 139 of *Proceedings of Machine Learning Research*, pp. 8748–8763, PMLR.

[18] Yusong Wu*, Ke Chen*, Tianyu Zhang*, Yuchen Hui*, Taylor Berg-Kirkpatrick, and Shlomo Dubnov, "Large-scale contrastive language-audio pretraining with feature fusion and keyword-to-caption augmentation," in *IEEE International Conference on Acoustics, Speech and Signal Processing, ICASSP*, 2023.

[19] Shentong Mo, Jing Shi, and Yapeng Tian, "Text-to-audio generation synchronized with videos," 2024.

[20] Shawn Hershey, Sourish Chaudhuri, Daniel P. W. Ellis, Jort F. Gemmeke, Aren Jansen, R. Channing Moore, Manoj Plakal, Devin Platt, Rif A. Saurous, Bryan Seybold, Malcolm Slaney, Ron J. Weiss, and Kevin Wilson, "Cnn architectures for large-scale audio classification," in *2017 IEEE International Conference on Acoustics, Speech and Signal Processing (ICASSP)*, 2017, pp. 131–135.

[21] Tomas Mikolov, Kai Chen, Greg Corrado, and Jeffrey Dean, "Efficient estimation of word representations in vector space," 2013.

[22] Huan Liao, Haonan Han, Kai Yang, Tianjiao Du, Rui Yang, Qinmei Xu, Zunnan Xu, Jingquan Liu, Jiasheng Lu, and Xiu Li, "Baton: Aligning text-to-audio model using human preference feedback," in *Proceedings of the Thirty-Third International Joint Conference on Artificial Intelligence, IJCAI-24*, Kate Larson, Ed. 8 2024, pp. 4542–4550, International Joint Conferences on Artificial Intelligence Organization, Main Track.

[23] Heng Wang, Jianbo Ma, Santiago Pascual, Richard Cartwright, and Weidong Cai, "V2a-mapper: A lightweight solution for vision-to-audio generation by connecting foundation models," in *Proceedings of the AAAI Conference on Artificial Intelligence*, 2024.

[24] Simian Luo, Chuanhao Yan, Chenxu Hu, and Hang Zhao, "Diff-foley: Synchronized video-to-audio synthesis with latent diffusion models," 2023.

[25] Gehui Chen, Guan'an Wang, Xiaowen Huang, and Jitao Sang, "Semantically consistent video-to-audio generation using multimodal language large model," 2024.

[26] Yiming Zhang, Yicheng Gu, Yanhong Zeng, Zhening Xing, Yuancheng Wang, Zhizheng Wu, and Kai Chen, "Foleycrafter: Bring silent videos to life with lifelike and synchronized sounds," 2024.

[27] Zehan Wang, Yang Zhao, Xize Cheng, Haifeng Huang, Jiageng Liu, Li Tang, Linjun Li, Yongqi Wang, Aoxiong Yin, Ziang Zhang, and Zhou Zhao, "Connecting multi-modal contrastive representations," 2023.

[28] Fei Zhao, Chengcui Zhang, and Baocheng Geng, "Deep multimodal data fusion," *ACM Comput. Surv.*, vol. 56, no. 9, apr 2024.

[29] Xinhao Mei, Varun Nagaraja, Gael Le Lan, Zhaoheng Ni, Ernie Chang, Yangyang Shi, and Vikas Chandra, "Foleygen: Visually-guided audio generation," 2023.

[30] Yazhou Xing, Yingqing He, Zeyue Tian, Xintao Wang, and Qifeng Chen, "Seeing and hearing: Open-domain visual-audio generation with diffusion latent aligners," in *CVPR*, 2024.

[31] Zhen Qin, Daoyuan Chen, Wenhao Zhang, Liuyi Yao, Yilun Huang, Bolin Ding, Yaliang Li, and Shuiguang Deng, "The synergy between data and multi-modal large language models: A survey from co-development perspective," 2024.

[32] Dongyao Zhu, Bowen Lei, Jie Zhang, Yanbo Fang, Yiqun Xie, Ruqi Zhang, and Dongkuan Xu, "Rethinking data distillation: Do not overlook calibration," in *Proceedings of the IEEE/CVF International Conference on Computer Vision (ICCV)*, October 2023, pp. 4935–4945.

[33] Meike Zehlike, Ke Yang, and Julia Stoyanovich, "Fairness in ranking, part ii: Learning-to-rank and recommender systems," *ACM Comput. Surv.*, vol. 55, no. 6, dec 2022.

[34] Zehan Wang, Ziang Zhang, Luping Liu, Yang Zhao, Haifeng Huang, Tao Jin, and Zhou Zhao, "Extending multi-modal contrastive representations," 2023.

[35] Zehan Wang, Ziang Zhang, Xize Cheng, Rongjie Huang, Luping Liu, Zhenhui Ye, Haifeng Huang, Yang Zhao, Tao Jin, Peng Gao, and Zhou Zhao, "Freebind: Free lunch in unified multimodal space via knowledge fusion," 2024.

[36] Rohit Girdhar, Alaaeldin El-Nouby, Zhuang Liu, Mannat Singh, Kalyan Vasudev Alwala, Armand Joulin, and Ishan Misra, "Imagebind: One embedding space to bind them all," in *CVPR*, 2023.

[37] Bin Zhu, Bin Lin, Munan Ning, Yang Yan, Jiaxi Cui, Wang HongFa, Yatian Pang, Wenhao Jiang, Junwu Zhang, Zongwei Li, Cai Wan Zhang, Zhifeng Li, Wei Liu, and Li Yuan, "Languagebind: Extending video-language pretraining to n-modality by language-based semantic alignment," 2023.

[38] Yiyuan Zhang, Kaixiong Gong, Kaipeng Zhang, Hongsheng Li, Yu Qiao, Wanli Ouyang, and Xiangyu Yue, "Meta-transformer: A unified framework for multimodal learning," *arXiv preprint arXiv:2307.10802*, 2023.